\documentclass[11pt,a4paper]{article}
\usepackage[hyperref]{acl2020}
\usepackage{times}
\usepackage{latexsym}
\usepackage{amssymb}

\usepackage{microtype}

\aclfinalcopy %

\usepackage{xspace,mfirstuc,tabulary}
\usepackage{amsmath,bm}
\usepackage{multirow}
\usepackage{color}
\usepackage{xcolor}
\usepackage{multicol}
\usepackage{tabularx}
\usepackage{booktabs}
\usepackage{pbox}
\usepackage{framed}
\usepackage{array}
\usepackage{enumitem}
\usepackage{chngpage}
\usepackage{threeparttable}
\usepackage{dsfont}
\usepackage{graphicx}
\usepackage{enumitem}
\usepackage{amsfonts}

\newcommand{\eg}{{e.g.}}
\newcommand{\ie}{{i.e.}}
\newcommand{\vs}{{vs.}}
\newcommand{\hhide}[1]{}

\newcommand\blfootnote[1]{%
  \begingroup
  \renewcommand\thefootnote{}\footnote{#1}%
  \addtocounter{footnote}{-1}%
  \endgroup
}

\title{Dialogue-Based Relation Extraction}

\author{
  Dian Yu\textsuperscript{1}\textsuperscript{\dag}  ~~ Kai Sun\textsuperscript{2}\textsuperscript{\dag} ~~ Claire Cardie\textsuperscript{2} ~~ Dong Yu\textsuperscript{1}\\
 \textsuperscript{1}Tencent AI Lab, Bellevue, WA\\
 \textsuperscript{2}Cornell University, Ithaca, NY \\
   \{yudian, dyu\}@tencent.com, ks985@cornell.edu,  cardie@cs.cornell.edu \\
}

\date{}

\begin{document}
\maketitle                          

\blfootnote{$\dag$ Equal contribution.}

\begin{abstract}

We present the first human-annotated dialogue-based relation extraction (RE) dataset DialogRE, aiming to support the prediction of relation(s) between two arguments that appear in a dialogue. We further offer DialogRE as a platform for studying cross-sentence RE as most facts span multiple sentences. We argue that speaker-related information plays a critical role in the proposed task, based on an analysis of similarities and differences between dialogue-based and traditional RE tasks. Considering the timeliness of communication in a dialogue, we design a new metric to evaluate the performance of RE methods in a conversational setting and investigate the performance of several representative RE methods on DialogRE. Experimental results demonstrate that a speaker-aware extension on the best-performing model leads to gains in both the standard and conversational evaluation settings. DialogRE is available at \url{https://dataset.org/dialogre/}.

\end{abstract}

\section{Introduction}

Cross-sentence relation extraction, which aims to identify relations between two arguments that are not mentioned in the same sentence or relations that cannot be supported by any single sentence, is an essential step in building knowledge bases from large-scale corpora automatically~\cite{ji2010overview,swampillai-stevenson-2010-inter,surdeanu2013overview}. It has yet to receive extensive study in natural language processing, however. In particular, although dialogues readily exhibit cross-sentence relations, most existing relation extraction tasks focus on texts from formal genres such as professionally written and edited news reports or well-edited websites~\cite{elsahar2018t,yao2019docred,mesquita-etal-2019-knowledgenet,grishman2019twenty}, while dialogues have been under-studied.

In this paper, we take an initial step towards studying relation extraction in dialogues by constructing the first human-annotated dialogue-based relation extraction dataset, \textbf{DialogRE}. Specifically, we annotate all occurrences of $36$ possible relation types that exist between pairs of arguments in the 1,788 dialogues originating from the complete transcripts of \textit{Friends}, a corpus that has been widely employed in dialogue research in recent years~\cite{catizone-etal-2010-using,chen2016character,chen2017robust,zhou2018they,rashid-blanco-2018-characterizing,yang2019friendsqa}.  Altogether, we annotate 10,168 relational triples.
For each (\emph{subject, relation type, object}) triple, we also annotate the minimal contiguous text span that most clearly expresses the relation; this may enable researchers to explore relation extraction methods that provide fine-grained explanations along with evidence sentences. For example, the bolded text span \emph{``brother''} in Table~\ref{tab:sample1} indicates the \textsc{per:siblings} relation (R1 and R2) between speaker 2 (S2) and \emph{``Frank''}.  %

\begin{table}[t!]
\centering
\footnotesize
\begin{tabular}{p{0.2cm}lll}
\toprule
\textbf{S1}:    & \multicolumn{3}{p{6.5cm}}{Hey Pheebs.}                         \\
\textbf{S2}:    & \multicolumn{3}{p{6.5cm}}{Hey!}                         \\
\textbf{S1}:    & \multicolumn{3}{p{6.5cm}}{Any sign of your \textbf{brother}?}                         \\
\textbf{S2}:    & \multicolumn{3}{p{6.5cm}}{No, but he's always late.}                         \\
\textbf{S1}:    & \multicolumn{3}{p{6.5cm}}{I thought you only met him once?}                         \\
\textbf{S2}:    & \multicolumn{3}{p{6.5cm}}{Yeah, I did. I think it sounds y'know big sistery, y'know, `Frank's always late.'}                         \\
\textbf{S1}:    & \multicolumn{3}{p{6.5cm}}{Well relax, he'll be here.}                         \\
\midrule
               & \textbf{Argument pair}               & \textbf{Trigger}        &  \textbf{Relation type}   \\
\textbf{R1}    & (Frank, S2)          & brother       &  per:siblings               \\
\textbf{R2}    & (S2, Frank)          & brother       &  per:siblings               \\
\textbf{R3}    & (S2, Pheebs)         & none          &  per:alternate\_names               \\
\textbf{R4}    & (S1, Pheebs)         & none          &  unanswerable                 \\
\bottomrule
\end{tabular}
\caption{A dialogue and its associated instances in DialogRE. S1, S2: anoymized speaker of each utterance.}
\label{tab:sample1}
\end{table}

Our analysis of DialogRE indicates that the supporting text for most (approximately $96.0\%$) annotated relational triples includes content from multiple sentences, making the dataset ideal for studying cross-sentence relation extraction. This is perhaps because of the higher person pronoun frequency~\cite{biber1991variation} and lower information density~\cite{wang-liu-2011-summarization} in conversational texts than those in formal written texts. In addition, $65.9\%$ of relational triples involve arguments that never appear in the same turn, suggesting that multi-turn information may play an important role in dialogue-based relation extraction. For example, to justify that \emph{``Pheebs''} is an alternate name of S2 in Table~\ref{tab:sample1}, the response of S2 in the second turn is required as well as the first turn.
We next conduct a thorough investigation of the similarities and differences between dialogue-based and traditional relation extraction tasks by comparing DialogRE and the Slot Filling dataset~\cite{mcnamee2009overview,ji2010overview,ji2011overview,surdeanu2013overview,surdeanu2014overview}, and we argue that a relation extraction system should be aware of speakers in dialogues. In particular, most relational triples in DialogRE ($89.9\%$) signify either an attribute of a speaker or a relation between two speakers. The same phenomenon occurs in an existing knowledge base constructed by encyclopedia collaborators, relevant to the same dialogue corpus we use for annotation (Section~\ref{sec:comparison_with_kb}). Unfortunately, most previous work directly applies existing relation extraction systems to dialogues without explicitly considering the speakers involved~\cite{yoshino-2011-spoken,wang-cardie-2012}.

Moreover, traditional relation extraction methods typically output a set of relations only after they have read the entire document and are free to rely on the existence of multiple mentions of a relation throughout the text to confirm its existence. However, these methods may be insufficient for powering a number of practical real-time dialogue-based applications such as chatbots, which would likely require recognition of a relation at its first mention in an interactive conversation. To encourage automated methods to identify the relationship between two arguments in a dialogue as early as possible, we further design a new performance evaluation metric for the conversational setting, which can be used as a supplement to the standard F$1$ measure (Section~\ref{sec:evaluation_metrics}). 

In addition to dataset creation and metric design, we adapt a number of strong, representative learning-based relation extraction methods~\cite{zeng2014relation,cai2016bidirectional,yao2019docred,bert2018} and evaluate them on DialogRE to establish baseline results on the dataset going forward. We also extend the best-performing method~\cite{bert2018} among them by letting the model be aware of the existence of arguments that are dialogue participants (Section~\ref{sec:sec:baselines}). Experiments on DialogRE demonstrate that this simple extension nevertheless yields substantial gains on both standard and conversational RE evaluation metrics, supporting our assumption regarding the critical role of tracking speakers in dialogue-based relation extraction (Section~\ref{sec:experiment}).

The primary contributions of this work are as follows: (\textbf{i}) we construct the first human-annotated dialogue-based relation extraction dataset and thoroughly investigate the similarities and differences between dialogue-based and traditional relation extraction tasks, (\textbf{ii}) we design a new conversational evaluation metric that features the timeliness aspect of interactive communications in dialogue, and (\textbf{iii}) we establish a set of baseline relation extraction results on DialogRE using standard learning-based techniques and further demonstrate the importance of explicit recognition of speaker arguments in dialogue-based relation extraction.%

\section{Data Construction}
\label{sec:construction}
We use the transcripts of all ten seasons ($263$ episodes in total) of an American television situation comedy \emph{Friends}, covering a range of topics. We remove all content (usually in parentheses or square brackets) that describes non-verbal information such as behaviors and scene information. %

\begin{table*}[ht!]
\centering
\footnotesize
\begin{tabular}{llllll}
\toprule
\bf ID & \bf Subject & \bf{Relation Type}  & \bf Object     & \bf Inverse Relation  & \bf TR (\%)\\
\midrule
1 & PER  & per:positive\_impression             & NAME      &                       & 70.4  \\
2 & PER  & per:negative\_impression             & NAME      &                       & 60.9 \\ %
3 & PER  & per:acquaintance                     & NAME      & per:acquaintance      & 22.2 \\
4 & PER  & per:alumni                           & NAME      & per:alumni            & 72.5 \\
5 & PER  & per:boss                             & NAME      & per:subordinate       & 58.1 \\
6 & PER  & per:subordinate                      & NAME      & per:boss              & 58.1 \\
7 & PER  & per:client                           & NAME      &                       & 50.0 \\
8 & PER  & per:dates                            & NAME      & per:dates             & 72.5 \\
9 & PER  & per:friends                          & NAME      & per:friends           & 94.7 \\
10 & PER  & per:girl/boyfriend                   & NAME     & per:girl/boyfriend    & 86.1 \\
11 & PER  & per:neighbor                         & NAME     & per:neighbor          & 71.2 \\
12 & PER  & per:roommate                         & NAME     & per:roommate          & 89.9\\
13 & PER  & per:children$^\star$                 & NAME     & per:parents           & 85.4\\
14 & PER  & per:other\_family$^\star$            & NAME     & per:other\_family     & 52.0\\
15 & PER  & per:parents$^\star$                  & NAME     & per:children          & 85.4\\
16 & PER  & per:siblings$^\star$                 & NAME     & per:siblings          & 80.5\\
17 & PER  & per:spouse$^\star$                   & NAME     & per:spouse            & 86.7\\
18 & PER  & per:place\_of\_residence$^{\star\star}$             & NAME     &  gpe:residents\_of\_place    & 42.9  \\
19 & PER  & per:place\_of\_birth$^{\star\star}$                  & NAME     &  gpe:births\_in\_place      & 100.0 \\
20 & PER  & per:visited\_place                   & NAME     &   gpe:visitors\_of\_place                   & 43.0 \\
21 & PER  & per:origin$^\star$                   & NAME     &                                             & 3.8  \\    
22 & PER  & per:employee\_or\_member\_of$^\star$ & NAME     & org:employees\_or\_members                  & 47.2 \\
23 & PER  & per:schools\_attended$^\star$        & NAME     &   org:students                              & 37.5 \\
24 & PER  & per:works                            & NAME     &                                             & 27.0 \\
25 & PER  & per:age$^\star$                      & VALUE    &                                             & 0.0  \\
26 & PER  & per:date\_of\_birth$^\star$          & VALUE    &                                             & 66.7 \\
27 & PER  & per:major                            & STRING   &                                             & 50.0 \\
28 & PER  & per:place\_of\_work                  & STRING   &                                             & 45.1 \\ %
29 & PER  & per:title$^\star$                    & STRING   &                                             & 0.5  \\
30 & PER  & per:alternate\_names$^\star$         & NAME/STRING     &                                      & 0.7  \\
31 & PER  & per:pet                              & NAME/STRING     &                                      & 0.3  \\
32 & GPE  & gpe:residents\_of\_place$^{\star\star}$ & NAME     &   per:place\_of\_residence               & 42.9 \\
33 & GPE  & gpe:births\_in\_place$^{\star\star}$    & NAME     &   per:place\_of\_birth                   & 100.0 \\
34 & GPE  & gpe:visitors\_of\_place       & NAME     &   per:visited\_place                               & 43.0   \\
35 & ORG  & org:employees\_or\_members     & NAME     &   per:employee\_or\_member\_of                    & 47.2\\
36 & ORG  & org:students$^\star$          & NAME          &   per:schools\_attended                       & 37.5\\
37 & NAME & unanswerable                  & NAME/STRING/VALUE   &                                         & ---\\
\bottomrule
\end{tabular}
\caption{Relation Types in DialogRE. Relation types with $\star$ represent the existing relation types defined in the TAC-KBP SF task, and we combine three SF fine-grained relation types about cities, states, and countries in a single relation type with $\star\star$. TR: Trigger ratio, representing the percentage of relational triples of a certain relation type that are accompanied by triggers.}
\label{tab:data:stat}
\end{table*}

\subsection{Relation Schema}
\label{sec:sec:rel_schema}
We follow the slot descriptions\footnote{http://surdeanu.info/kbp2014/def.php.} of the Slot Filling (SF) task in the Text Analysis Conference Knowledge Base Population (TAC-KBP)~\cite{mcnamee2009overview,ji2010overview,ji2011overview,surdeanu2013overview,surdeanu2014overview}, which primarily focuses on biographical attributes of person (PER) entities and important attributes of organization (ORG) entities. As the range of topics in \emph{Friends} is relatively restricted compared to large-scale news corpora such as Gigaword~\cite{parker2011english}, some relation types (\eg, \textsc{per:charges}, and \textsc{org:subsidiaries}) seldom appear in the texts. Additionally, we consider new relation types such as \textsc{per:girl/boyfriend} and \textsc{per:neighbor} that frequently appear in \emph{Friends}. We list all $36$ relation types that have at least one relational instance in the transcripts in Table~\ref{tab:data:stat} and provide definitions and examples of new relation types in Appendix~\ref{sec:newreldef}. %

\subsection{Annotation}
\label{sec:sec:annotation}

We focus on the annotation of \emph{relational triples} (\ie, (\emph{subject, relation type, object})) in which at least one of the arguments is a named entity. We regard an uninterrupted stream of speech from one speaker and the name of this speaker as a \emph{turn}. %

As we follow the TAC-KBP guideline to annotate relation types and design new types, we use internal annotators (two authors of this paper) who are familiar with this task. For a pilot annotation, annotator A annotates relational triples in each scene in all transcripts and form a \emph{dialogue} by extracting the shortest snippet of contiguous turns that covers all annotated relational triples and sufficient supportive contexts in this scene. The guidelines are adjusted during the annotation.\footnote{As the pilot annotation only involves one annotator, we admit there may exist a certain degree of bias in defining new relation types and labeling argument pairs.} We prefer to use \emph{speaker name} (\ie, the first word or phrase of a turn, followed by a colon) as one argument of a speaker-related triple if the corresponding full names or alternate names of the speaker name also appear in the same dialogue, except for relation \textsc{per:alternate\_names} in which both mentions should be regarded as arguments. %
For an \emph{argument pair} (\ie, (\emph{subject, object})), there may exist multiple relations between them, and we annotate all instances of all of them. For each triple, we also annotate its \emph{trigger}: the smallest extent (\ie, span) of contiguous text in the dialogue that most clearly indicates the existence of the relation between two arguments. If there exist multiple spans that can serve as triggers, we only keep one for each triple. For relation types such as \textsc{per:title} and \textsc{per:alternate\_names}, it is difficult to identify such supportive contexts, and therefore we leave their triggers empty. For each relational triple, we annotate its inverse triple if its corresponding inverse relation type exists in the schema (\eg, \textsc{per:children} and \textsc{per:parents}) while the trigger remains unchanged.  %

In the second process, annotator B annotates the possible relations between candidate pairs annotated by annotator A (previous relation labels are hidden). Cohen's kappa among annotators is around $0.87$. We remove the cases when annotators cannot reach a consensus. On average, each dialogue in DialogRE contains $4.5$ relational triples and $12.9$ turns, as shown in Table~\ref{tab:statistics}. See Table~\ref{tab:sample1} for relational triple examples (R1, R2, and R3).

\begin{table}[h!]
\centering
\footnotesize
\begin{tabular}{ll}
\toprule
\textbf{DialogRE}    & \\
\midrule
Average dialogue length (in tokens)  &  225.8  \\
Average \# of turns                  &  12.9   \\
Average \# of speakers               &  3.3    \\
Average \# of sentences              &  21.8  \\
Average \# of relational instances   &  4.5     \\
Average \# of no-relation instances  &  1.2   \\
\bottomrule
\end{tabular}
\caption{Statistics per dialogue of DialogRE.}
\label{tab:statistics}
\end{table}

\subsection{Negative Instance Generation, Data Split, and Speaker Name Anonymization}

After our first round of annotation, we use any two annotated arguments associated with each dialogue to generate candidate relational triples, in which the relation between two arguments is unanswerable based on the given dialogue or beyond our relation schema. We manually filter out candidate triples for which there is ``obviously'' no relation between an argument pair in consideration of aspects such as argument type constraints (\eg, relation~\textsc{per:schools\_attended} can only exist between a PER name and an ORG name). After filtering, we keep 2,100 triples in total, whose two arguments are in ``no relation'', and we finally have 10,168 triples for 1,788 dialogues. We randomly split them at the dialogue level, with $60\%$ for training, $20\%$ for development, and $20\%$ for testing.

The focus of the proposed task is to identify relations between argument pairs based on a dialogue, rather than exploiting information in DialogRE beyond the given dialogue or leveraging external knowledge to predict the relations between arguments (\eg, characters) specific to a particular television show. Therefore, we anonymize all speaker names (Section~\ref{sec:sec:annotation}) in each dialogue and annotated triples and rename them in chronological order within the given dialogue. For example, S$1$ and S$2$ in Table~\ref{tab:sample1} represent the original speaker names \emph{``Rachel''} and \emph{``Phoebe''}, respectively.

\section{Data Comparisons and Discussions}
\label{sec:data_analysis}

\subsection{Comparison Between DialogRE and SF}
\label{sec:comparison_with_sf}

\begin{figure}[h!]
   \begin{center}
   \includegraphics[width=0.46\textwidth]{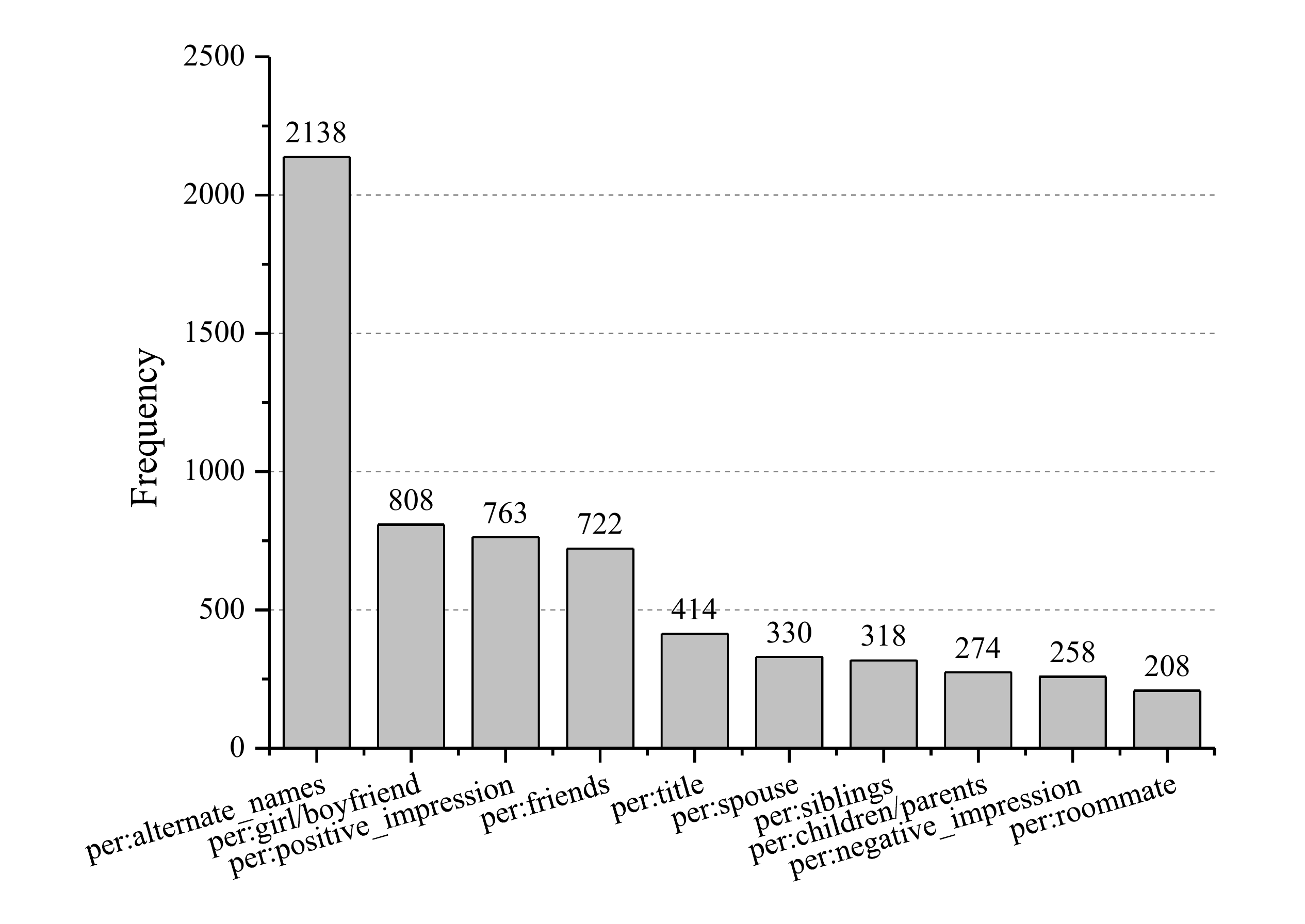}
   \end{center}
 \caption{Relation type distribution in DialogRE.}
 \label{fig:typedre}
\end{figure}

\begin{figure}[h!]
   \begin{center}
   \includegraphics[width=0.46\textwidth]{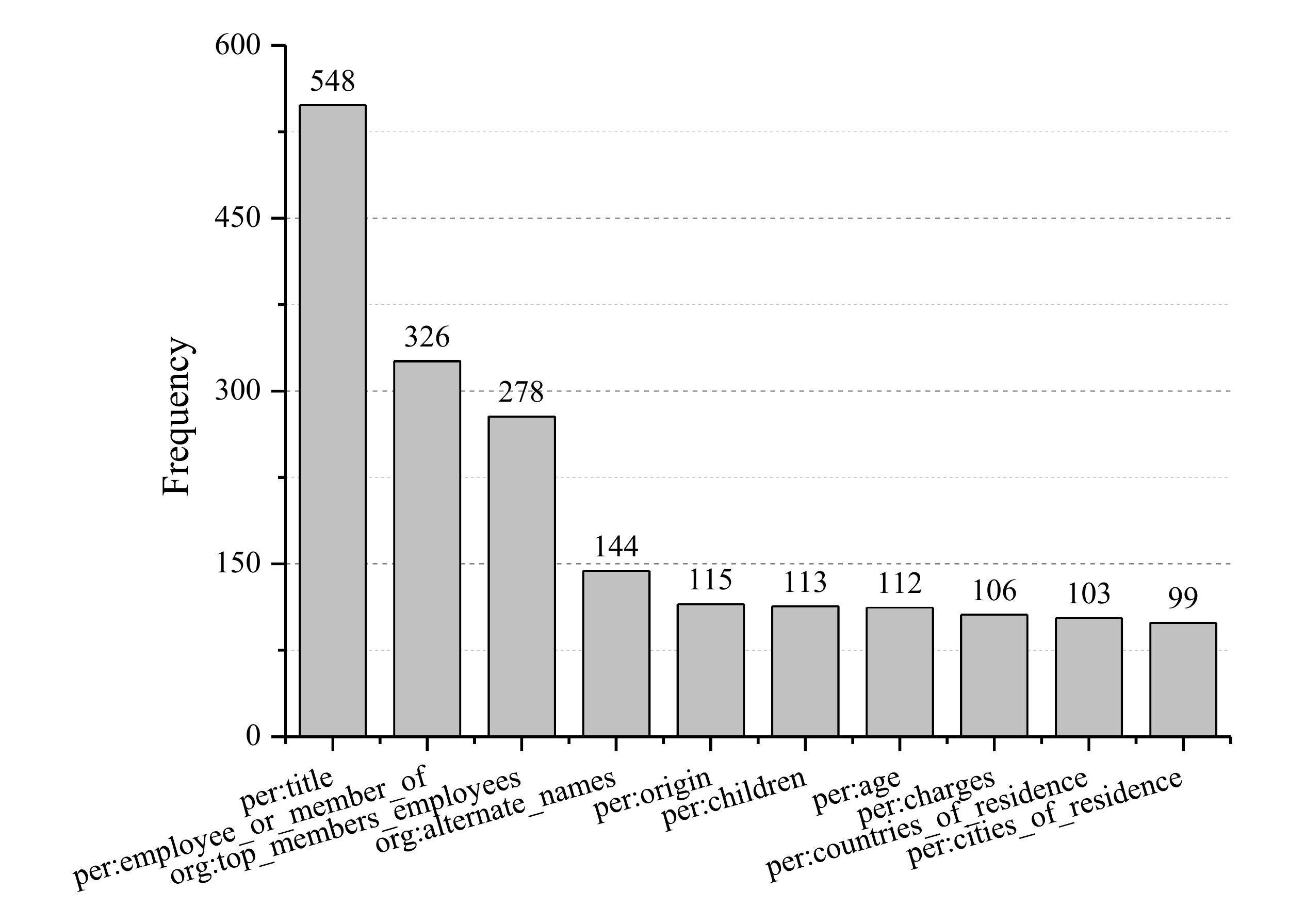}
   \end{center}
\caption{Relation type distribution in SF (2013-2014).} 
 \label{fig:typesf}
\end{figure}

As a pilot study, we examine the similarities and differences between dialogue-based and traditional relation extraction datasets that are manually annotated. We compare DialogRE with the official SF (2013-2014) dataset~\cite{surdeanu2013overview,surdeanu2014overview} as $47.2\%$ of relation types in DialogRE originate from the SF relation types (Section~\ref{sec:sec:rel_schema}), and $92.2\%$ of the source documents in it that contain ground truth relational triples are formally written newswire reports ($72.8\%$) or well-edited web documents ($19.4\%$) compared to the remaining documents from discussion fora. 
We show the relation distributions in DialogRE and SF in Figure~\ref{fig:typedre} and Figure~\ref{fig:typesf} (Appendix~\ref{sec:reltypedist}), respectively. Half of the top ten relation types in DialogRE are newly defined (\textsc{per:girl/boyfriend}, \textsc{per:positive(negative)\_impression}, \textsc{per:friends}, and \textsc{per:roommate}), partially justifying the need for new relation types.

\medskip

\noindent \textbf{Argument Type}: Based on the predefined SF and DialogRE relation types, a subject is expected to be an entity of type PER, ORG, or geo-political entity (GPE). Notably, subjects of most relational triples ($96.8\%$ \vs~ $69.7\%$ in the SF dataset) in DialogRE are person names. The coarse-grained object type is entity, string, or value (\ie, a numerical value or a date). As shown in Table~\ref{tab:object_type}, we observe that a higher proportion ($80.1\%$) of objects are entities in DialogRE compared to that in SF ($65.3\%$).

\begin{table}[h!]
\footnotesize
\centering
\begin{tabular}{lll}
\toprule
                         & \textbf{DialogRE}              & \textbf{SF} \\
\midrule
Entity                   & 80.1  (6,460)                  & 65.3 (2,167)\\
String                   & 18.9  (1,524)                  & 25.4 (843)\\
Value                    & 1.0   (84)                     & 9.2 (306) \\
\bottomrule
\end{tabular}
\caption{Coarse-grained object type distributions (\%) of DialogRE and SF with frequencies in brackets.} 
\label{tab:object_type}
\end{table}

In particular, the subjects of $77.3\%$ of relational triples are speaker names, and more than $90.0\%$ of relational triples contain at least one speaker argument. The high percentage of ``speaker-centric'' relational triples and the low percentage of ORG and GPE arguments in DialogRE is perhaps because the transcripts for annotation are from a single situation comedy that involves a small group of characters in a very limited number of scenes (see more discussions in Section~\ref{sec:limitation}).

\medskip
\noindent \textbf{Distance Between Argument Pairs}: It has been shown that there is a longer distance between two arguments in the SF dataset~\cite{surdeanu2013overview,huang-2017-improving} compared to that in many widely used human-annotated relation extraction datasets such as ACE~\cite{doddington2004automatic} and SemEval~\cite{hendrickx2010semeval}. However, it is not trivial to compute an accurate distance between two arguments in a dialogue, especially for cases containing arguments that are speaker names. We instead consider different types of distances (\eg, average and minimum) between two argument mentions in a dialogue. We argue that DialogRE exhibits a similar level of difficulty as SF from the perspective of the distance between two arguments. $41.3\%$ of arguments are separated by at least seven words even considering the minimum distance, and the percentage can reach as high as $96.5\%$ considering the average distance, contrast with $46.0$\% in SF~\cite{huang-2017-improving} and $59.8\%$ in a recently released cross-sentence relation extraction dataset DocRED, in which Wikipedia articles serve as documents~\cite{yao2019docred}. Note that the provenance/evidence sentences in SF and DocRED are provided by automated systems or annotators. Also, $95.6\%$ of relational triples from an annotated subset of DialogRE (Section~\ref{sec:sec:results}) require reasoning over multiple sentences in a dialogue, compared with $40.7\%$ in DocRED (Table~\ref{tab:cross-sentence-re-datasets}). See Figure~\ref{fig:distance} in Appendix~\ref{sec:argdist} for more details.

\subsection{Comparison Between DialogRE and Existing Relational Triples}
\label{sec:comparison_with_kb}
We also collect 2,330 relational triples related to \emph{Friends}, which are summarized by a community of contributors, from a collaborative encyclopedia.\footnote{https://friends.fandom.com/wiki/Friends.} We remove triples of content-independent relation types such as \textsc{directed\_by}, \textsc{guest\_stars}, and \textsc{number\_of\_episodes}. 

We find that $94.5\%$ of all $237$ relation types in these triples can be mapped to one of the $36$ relation types in our relation schema (\eg, \textsc{husband}, \textsc{ex-husband}, and \textsc{wife} can be mapped to \textsc{per:spouse}) except for the remaining relatively rare or implicit relation types such as \textsc{prom\_date}, \textsc{gender}, and \textsc{kissed}, demonstrating the relation schema we use for annotation is capable of covering most of the important relation types labeled by the encyclopedia community of contributors. 

On the other hand, the relatively small number of the existing triples and the moderate size of our annotated triples in DialogRE may suggest the low information density~\cite{wang-liu-2011-summarization} in conversational speech in terms of relation extraction. For example, the average annotated triple per sentence in DialogRE is merely $0.21$, compared to other exhaustively annotated datasets ACE ($0.73$) and KnowledgeNet~\cite{mesquita-etal-2019-knowledgenet} ($1.44$), in which corpora are formal written news reports and Wikipedia articles, respectively.

\subsection{Discussions on Triggers}
As annotated triggers are rarely available in existing relation extraction datasets~\cite{aguilar-2014-comparison}, the connections between different relation types and trigger existence are under-investigated. 

\medskip

\noindent \textbf{Relation Type}: In DialogRE, $49.6\%$ of all relational triples are annotated with triggers. We find that argument pairs are frequently accompanied by triggers when (1) arguments have the same type such as \textsc{per:friends}, (2) strong emotions are involved (\eg, \textsc{per:positive(negative)\_impression}), or (3) the relation type is related to death or birth (\eg, \textsc{gpe:births\_in\_place}). In comparison, a relation between two arguments of different types (\eg, \textsc{per:origin} and \textsc{per:age}) is more likely to be implicitly expressed instead of relying on triggers. This is perhaps because there exist fewer possible relations between such an argument pair compared to arguments of the same type, and a relatively short distance between such an argument pair might be sufficient to help the listeners understand the message correctly. For each relation type, we report the percentage of relational triples with triggers in Table~\ref{tab:data:stat}. 

\medskip

\noindent \textbf{Argument Distance}: We assume the existence of triggers may allow a longer distance between argument pairs in a text as they help to decrease ambiguity. This assumption may be empirically validated by the longer average distance ($68.3$ tokens) between argument pairs with triggers in a dialogue, compared to the distance ($61.2$ tokens) between argument pairs without any triggers. %

\section{Task Formulations and Methods}
\label{sec:task}

\subsection{Dialogue-Based Relation Extraction}
\label{sec:evaluation_metrics}

Given a dialogue $D=s_1:t_1,s_2:t_2,\ldots,s_m:t_m$ and an argument pair $(a_1, a_2)$, where $s_i$ and $t_i$ denote the speaker ID and text of the $i^{\text{th}}$ turn, respectively, and $m$ is the total number of turns, we evaluate the performance of approaches in extracting relations between $a_1$ and $a_2$ that appear in $D$ in the following two settings.

\medskip

\noindent \textbf{Standard Setting}: As the standard setting of relation extraction tasks, we regard dialogue $D$ as document $d$. The input is $a_1$, $a_2$, and $d$, and the expected output is the relation type(s) between $a_1$ and $a_2$ based on $d$. We adopt F$1$, which is the harmonic mean of precision (P) and recall (R), for evaluation.

\medskip

\noindent \textbf{Conversational Setting}: 
Instead of only considering the entire dialogue, here we can regard the first $i\leq m$ turns of the dialogue as $d$. Accordingly, we propose a new metric $\text{F}1_\text{c}$, the harmonic mean of conversational precision ($\text{P}_\text{c}$) and recall ($\text{R}_\text{c}$), as a supplement to the standard F$1$. We start by introducing some notation that will be used in the definition of $\text{F}1_\text{c}$. Let $O_{i}$ denote the set of predicted relation types when the input is $a_1$, $a_2$, and the first $i$ turns (\ie, $d=s_1:t_1,s_2:t_2,\ldots,s_i:t_i$). For an argument pair ($a_1$, $a_2$), let $L$ denote its corresponding set of relation types that are manually annotated based on the full dialogue. $R$ represents the set of $36$ relation types. By definition, $O_{i}, L\subseteq R$. We define that auxiliary function $\jmath(x)$ returns $m$ if $x$ does not appear in $D$. Otherwise, it returns the index of the turn where $x$ first appears.

We define auxiliary function $\imath(r)$ as: (i) For each relation type $r\in L$, if there exists an annotated trigger for $r$, $\imath(r)=\jmath(\lambda_r)$ where $\lambda_r$ denotes the trigger. Otherwise, $\imath(r)=m$. (ii) For each $r\in R\backslash L$, $\imath(r)=1$. We define the set of relation types that are evaluable based on the first $i$ turns by $E_i$: %
\begin{equation}
E_i=\left\{r\,|\,i\geq \max\{\jmath(a_1), \jmath(a_2), \imath(r)\}\right\}
\label{eq:ei}
\end{equation}
The interpretation of Equation~\ref{eq:ei} is that given $d$ containing the first $i$ turns in a dialogue, relation type $r$ associated with $a_1$ and $a_2$ is evaluable if $a_1$, $a_2$, and the trigger for $r$ have all been mentioned in $d$. 
The definition is based on our assumption that we can roughly estimate how many turns we require to predict the relations between two arguments based on the positions of the arguments and triggers, which most clearly express relations. See Section~\ref{sec:sec:results} for more discussions.

The conversational precision and recall for an input instance $D$, $a_1$, and $a_2$ are defined as:
\begin{equation}
\text{P}_\text{c}(D,a_1,a_2)=\frac{\sum_{i=1}^{m}|O_i\cap L\cap E_i|}{\sum_{i=1}^{m}|O_i\cap E_i|}
\end{equation}
\begin{equation}
\text{R}_\text{c}(D,a_1,a_2)=\frac{\sum_{i=1}^{m}|O_i\cap L\cap E_i|}{\sum_{i=1}^{m}|L\cap E_i|}
\end{equation}
We average the conversational precision/recall scores of all instances to obtain the final conversational precision/recall.
\begin{equation}
\text{P}_\text{c} = \frac{\sum_{D',a_1',a_2'}\text{P}_\text{c}(D',a_1',a_2')}{\sum_{D',a_1',a_2'}1}
\end{equation}
\begin{equation}
\text{R}_\text{c} = \frac{\sum_{D',a_1',a_2'}\text{R}_\text{c}(D',a_1',a_2')}{\sum_{D',a_1',a_2'}1}
\end{equation}
and $\text{F}1_\text{c}=2\cdot \text{P}_\text{c} \cdot \text{R}_\text{c}/(\text{P}_\text{c}+\text{R}_\text{c})$.

\begin{table*}[t]
\centering
\footnotesize
\begin{tabular}{lcc|cc}
\toprule
\multirow{2}{*}{\bf Method}          & \multicolumn{2}{c|}{\bf Dev} & \multicolumn{2}{c}{\bf Test} \\
                                      & \bf F$1$ ($\sigma$)      & \bf $\text{F}1_\text{c}$ ($\sigma$)  & \bf F$1$ ($\sigma$)  & \bf $\text{F}1_\text{c}$ ($\sigma$)   \\
\midrule
Majority   & 38.9 (0.0) &  38.7 (0.0) &  35.8 (0.0) & 35.8 (0.0) \\
CNN          & 46.1 (0.7)  & 43.7 (0.5)    & 48.0 (1.5)   & 45.0 (1.4)  \\
LSTM         & 46.7 (1.1)  & 44.2 (0.8)     & 47.4 (0.6)  & 44.9 (0.7)    \\
BiLSTM      &  48.1 (1.0)   & 44.3 (1.3)     & 48.6 (1.0)  & 45.0 (1.3)    \\
BERT &    60.6 (1.2)  & 55.4 (0.9)      & 58.5 (2.0)  & 53.2 (1.6) \\  
$\text{BERT}_{\text{S}}$  &  63.0 (1.5)    & 57.3 (1.2)    & 61.2 (0.9) & 55.4 (0.9) \\ 
\bottomrule
\end{tabular}
\caption{Performance of relation extraction methods on DialogRE in both the standard and conversational settings.} %
\label{tab:big_table}
\end{table*}

\subsection{Baselines}
\label{sec:sec:baselines}

\noindent \textbf{Majority}: If a given argument pair does not appear in the training set, output the majority relation type in the training set as the prediction. Otherwise, output the most frequent relation type associated with the two arguments in the training set.

\medskip

\noindent \textbf{CNN, LSTM, and BiLSTM}: Following previous work~\cite{yao2019docred}, we adapt three baselines~\cite{zeng2014relation,cai2016bidirectional} that use different document encoders. We refer readers to~\newcite{yao2019docred} for more details.

\medskip

\noindent \textbf{BERT}: We follow the framework of fine-tuning a pre-trained language model on a downstream task~\cite{radfordimproving} and use BERT~\cite{bert2018} as the pre-trained model. We concatenate the given $d$ and $(a_1, a_2)$ with classification token \texttt{[CLS]} and separator token \texttt{[SEP]} in BERT as the input sequence  \texttt{[CLS]}$d$\texttt{[SEP]}$a_1$\texttt{[SEP]}$a_2$\texttt{[SEP]}. We denote the final hidden vector corresponding to \texttt{[CLS]} as $C\in \mathbb{R}^{H}$, where $H$ is the hidden size. For each relation type $i$, we introduce a vector $W_i\in \mathbb{R}^{H}$ and obtain the probability $P_i$ of the existence of $i$ between $a_1$ and $a_2$ based on $d$ by $P_i=\text{sigmoid}(CW_i^T)$. The cross-entropy loss is used.

\medskip

\noindent \textbf{$\text{BERT}_\text{S}$}:  We propose a modification to the input sequence of the above BERT baseline with two motivations: (1) help a model locate the start positions of relevant turns based on the arguments that are speaker names, and (2) prevent a model from overfitting to the training data. Formally, given an argument pair $(a_1, a_2)$ and its associated document $d=s_1:t_1,s_2:t_2,\ldots,s_n:t_n$, we construct  $\hat{d}=\hat{s}_1:t_1,\hat{s}_2:t_2,\ldots,\hat{s}_n:t_n$, where $\hat{s}_i$ is:

\begin{equation}
\hat{s}_i = 
    \begin{cases}
      \hfil \texttt{[S\textsubscript{1}]} & \text{if}\ s_i=a_1 \\
      \hfil \texttt{[S\textsubscript{2}]} & \text{if}\ s_i=a_2
      \\
      \hfil s_i & \text{otherwise}
    \end{cases}
\end{equation}
where \texttt{[S\textsubscript{1}]} and \texttt{[S\textsubscript{2}]} are two newly-introduced special tokens. In addition, we define $\hat{a}_k$ ($k\in \{1,2\}$) to be \texttt{[S\textsubscript{k}]} if $\exists i( s_i=a_k)$, and $a_k$ otherwise. The modified input sequence to BERT is \texttt{[CLS]}$\hat{d}$\texttt{[SEP]}$\hat{a}_1$\texttt{[SEP]}$\hat{a}_2$\texttt{[SEP]}. In Appendix~\ref{sec:input_sequence}, we investigate in three alternative input sequences. It is worth mentioning that a modification that does not disambiguate speaker arguments from other arguments performs substantially worse than the above speaker-aware modification.

\section{Experiment}
\label{sec:experiment}

\subsection{Implementation Details}

\noindent \textbf{CNN, LSTM, and BiLSTM Baselines}: %
The CNN/LSTM/BiLSTM encoder takes as features GloVe word embeddings~\cite{pennington2014glove}, mention embeddings, and type embeddings. We assign the same mention embedding to mentions of the same argument and obtain the type embeddings based on named entity types of the two arguments. We use spaCy\footnote{https://spacy.io/.} for entity typing.

\medskip

\noindent \textbf{Language Model Fine-Tuning}:
We use the uncased base model of BERT released by~\citet{bert2018}. We truncate a document when the input sequence length exceeds $512$ and fine-tune BERT using a batch size of $24$ and a learning rate of $3\times 10^{-5}$ for $20$ epochs. Other parameters remain unchanged. The embeddings of newly-introduced special tokens (\eg, \texttt{[S\textsubscript{1}]}) are initialized randomly. %

\subsection{Results and Discussions}
\label{sec:sec:results}

We report the performance of all baselines in both the standard and conversational settings in Table~\ref{tab:big_table}. We run each experiment five times and report the average F$1$ and $\text{F}1_\text{c}$ along with standard deviation ($\sigma$). The fine-tuned BERT method already outperform other baselines (\eg, BiLSTM that achieves $51.1\%$ in F$1$ on DocRED~\cite{yao2019docred}), and our speaker-aware extension to the BERT baseline further leads to $2.7\%$ and $2.2\%$ improvements in F$1$ and $\text{F}1_\text{c}$, respectively, on the test set of DialogRE, demonstrating the importance of tracking speakers in dialogue-based relation extraction.

\medskip
\noindent \textbf{Conversational Metric}: 
We randomly select $269$ and $256$ instances, which are associated with $50$ dialogues from each of the dev and test sets, respectively. For each of relational instances ($188$ in total) that are previously labeled with triggers in the subsets, annotator A labels the smallest turn $i^\ast$ such that the first $i^\ast$ turns contain sufficient information to justify a relation. The average distance between $i^\ast$ and our estimation $\max\{\jmath(a_1), \jmath(a_2), \imath(r)\}$ in Equation~(\ref{eq:ei}) (Section~\ref{sec:evaluation_metrics}) is only $0.9$ turn, supporting our hypothesis that the positions of arguments and triggers may be good indicators for estimating the minimum turns for humans to make predictions.

For convenience, we use BERT for the following discussions and comparisons.

\medskip

\noindent \textbf{Ground Truth Argument Types}: Methods in Table~\ref{tab:big_table} are not provided with ground truth argument types considering the unavailability of this kind of annotation in practical use. To study the impacts of argument types on DialogRE, we report the performance of four methods, each of which additionally takes as input the ground truth argument types as previous work~\cite{zhang2017position,yao2019docred}. We adopt the same baseline for a direct comparison except that the input sequence is changed.

In \textbf{Method 1}, we simply extend the original input sequence of $\text{BERT}$ (Section~\ref{sec:sec:baselines}) with newly-introduced special tokens that represent argument types. The input sequence is \texttt{[CLS]}$d$\texttt{[SEP]}$\tau_1 a_1$\texttt{[SEP]}$\tau_2 a_2$\texttt{[SEP]}, where $\tau_i$ is a special token representing the argument type of $a_i$ ($i\in \{1,2\}$). For example, given $a_1$ of type PER and $a_2$ of type STRING, $\tau_1$ is \texttt{[PER]} and $\tau_2$ is \texttt{[STRING]}. %
In \textbf{Method 2}, we extend the input sequence of $\text{BERT}_\text{S}$ with $\tau_i$ defined in Method 1 (\ie, \texttt{[CLS]}$\hat{d}$\texttt{[SEP]}$\tau_1\hat{a}_1$\texttt{[SEP]}$\tau_1\hat{a}_2$\texttt{[SEP]}).
We also follow the input sequence of previous single-sentence relation extraction methods~\cite{shi2019simple,joshi2019spanbert} and refer them as \textbf{Method 3} and \textbf{4}, respectively. We provide the implementation details in Appendix~\ref{sec:gtargtype}. As shown in Table~\ref{tab:gtargtype}, the best performance achieved by Method 2 is not superior to that of $\text{BERT}_\text{S}$, which does not leverage ground truth argument types. Therefore, we guess that ground truth argument types may only provide a limited, if at all positive, contribution to the performance on DialogRE.  %

\begin{table}[h!]
\centering
\footnotesize
\begin{tabular}{lcccc}
\toprule
 & \bf Method 1 & \bf Method 2 & \bf Method 3 & \bf Method 4  \\
 & \bf F$1$ ($\sigma$) & \bf F$1$ ($\sigma$) & \bf F$1$ ($\sigma$)  & \bf F$1$ ($\sigma$) \\
\midrule
\bf Dev    & 60.6 (0.4) & \textbf{62.9} (1.2) & 55.6 (2.4) & 61.9 (1.4) \\
\bf Test  & 59.1 (0.7) & \textbf{60.5} (1.9) & 52.3 (3.2) & 59.7 (0.6) \\ %
\bottomrule
\end{tabular}
\caption{Performance comparison of methods with considering the ground truth argument types.}

\label{tab:gtargtype}
\end{table}

\noindent \textbf{Ground Truth Triggers}: We investigate what performance would be ideally attainable if the model could identify all triggers correctly. We append the ground truth triggers to the input sequence on the baseline, and the F$1$ of this model is $74.9\%$, a $16.4\%$ absolute improvement compared to the BERT baseline. In particular, through the introduction of triggers, we observe a $22.9\%$ absolute improvement in F$1$ on relation types whose inverse relation types are themselves (\eg, \textsc{per:roommate} and \textsc{per:spouse}). These experimental results show the critical role of triggers in dialogue-based relation extraction. However, trigger identification is perhaps as difficult as relation extraction, and it is labor-intensive to annotate large-scale datasets with triggers. Future research may explore how to identify triggers based on a small amount of human-annotated triggers as seeds~\cite{bronstein2015seed,yu2016unsupervised}.

\begin{table*}[t]
\centering
\footnotesize
\begin{tabular}{llllll}
\toprule
\bf Task                 & \bf style/source of doc  & \bf \# rel   & \bf cross rate$^\circ$ & \bf \# doc  & \bf \# triples$^\bullet$\\
\midrule
\multicolumn{5}{c}{----- distant supervision -----}                                       \\
\midrule
\newcite{peng2017cross}          & written/PubMed        & 4  & 75.2     &  960,000               &  140,661            \\
DocRED~\cite{yao2019docred}      & written/Wikipedia     & 96 & n/a   &  101,873                & 881,298             \\
T-REx~\cite{elsahar2018t}        & written/Wikipedia     & 353 & n/a     &   3 million           & 11 million          \\

\midrule
\multicolumn{5}{c}{----- human annotation -----}                                       \\ 
\midrule
BC5CDR~\cite{li2016biocreative}                      & written/PubMed                    &  1   & n/a    &  1,500                                  & 2,434             \\
DocRED~\cite{yao2019docred}                          & written/Wikipedia                 &  96  &  40.7   &  5,053                             & 56,354            \\
KnowledgeNet~\cite{mesquita-etal-2019-knowledgenet}  & written/Wikipedia and others     &  15  & n/a    &  4,991                  & 13,425            \\
\textbf{DialogRE} (this work)                        & \textbf{conversational}/Friends            &  36  & \bf 95.6    &  1,788                              & 8,068    \\
\bottomrule
\end{tabular}
\caption{\label{tab:related}Statistics of publicly available cross-sentence relation extraction datasets ($\circ$: the percentage (\%) of relational triples involving multiple sentences; $\bullet$: not include no-relation argument pairs).}
\label{tab:cross-sentence-re-datasets}
\end{table*}

\subsection{Error Analysis and Limitations}
\label{sec:limitation}
We analyze the outputs on the dev set and find that $\text{BERT}$ tends to make more mistakes when there exists an asymmetric inverse relation of the relation to be predicted compared to those that have symmetric inverse relations. For example, the baseline mistakenly predicts S2 as the subordinate of S1 based on the following dialogue:
\emph{``$\dots$S2: Oh. Well, I wish I could say no, but you can't stay \textbf{my assistant} forever. Neither can you Sophie, but for different reasons.
S1: God, I am so glad you don't have a problem with this, because if you did, I wouldn't even consider applying$\dots$"}. Introducing triggers into the input sequence leads to a relatively small gain ($11.0\%$ in F$1$ on all types with an asymmetric inverse relation) perhaps because inverse relation types share the same triggers (\eg, \emph{``my assistant''} serves as the trigger for both \textsc{per:boss} and \textsc{per:subordinate}). One possible solution may be the use of directed syntactic graphs constructed from the given dialogue, though the performance of coreference resolution and dependency parsing in dialogues may be relatively unsatisfying.  

A major limitation in DialogRE is that all transcripts for annotation are from \emph{Friends}, which may limit the diversity of scenarios and generality of the relation distributions. 
It may be useful to leverage existing triples in knowledge bases (\eg, \emph{Fandom}) for thousands of movies or TV shows using distant supervision~\cite{Mintz-ds-2009}, considering the time-consuming manual annotation process. In addition, dialogues in \emph{Friends} presents less variation based on linguistic features~\cite{biber1991variation} than natural conversations; nonetheless, compared to other registers such as personal letters and prepared speeches, there are noticeable linguistic similarities between natural conversations and television dialogues in \emph{Friends} ~\cite{quaglio2009television}.

\section{Related Work}

\textbf{Cross-Sentence Relation Extraction Datasets} Different from the sentence-level relation extraction (RE) datasets~\cite{roth2004linear,hendrickx2010semeval,riedel2010modeling,zhang2015relation,zhang2017position,han2018fewrel}, in which relations are between two arguments in the same sentence, we focus on cross-sentence RE tasks~\cite{ji2011overview,surdeanu2013overview,surdeanu2014overview} and present the first dialogue-based RE dataset, in which dialogues serve as input contexts instead of formally written sentences or documents. We compare DialogRE and existing cross-sentence RE datasets~\cite{li2016biocreative,quirk2017distant,yao2019docred,mesquita-etal-2019-knowledgenet} in Table~\ref{tab:cross-sentence-re-datasets}. In this paper, we do not consider relations that take relations or events as arguments and are also likely to span multiple sentences~\cite{pustejovsky-verhagen-2009-semeval,do-etal-2012-joint,moschitti-2013-long}.

\medskip

\noindent \textbf{Relation Extraction Approaches} Over the past few years, neural models have achieved remarkable success in RE~\cite{nguyen-grishman-2015-re,nguyen2015combining,adel-2016-comparing,yin2017comparative,levy2017zero,su2017global,song2018n,luo-2019-semi}, in which the input representation usually comes from shallow neural networks over pre-trained word and character embeddings~\cite{xu2015classifying,zeng2015distant,lin2016neural}. Deep contextualized word representations such as the ELMo~\cite{Peters:2018} are also applied as additional input features to boost the performance~\cite{luan2018multi}. A recent thread is to fine-tune pre-trained deep language models on downstream tasks~\cite{radfordimproving,bert2018}, leading to further performance gains on many RE tasks~\cite{alt2019improving,shi2019simple,baldini2019matching,peters-etal-2019-knowledge,wadden-etal-2019-entity}. We propose an improved method that explicitly considers speaker arguments, which are seldom investigated in previous RE methods.

\medskip

\noindent \textbf{Dialogue-Based Natural Language Understanding} To advance progress in spoken language understanding, researchers have studied dialogue-based tasks such as argument extraction~\cite{swanson-2015-argument}, named entity recognition~\cite{chen2016character,choi2018semeval,bowden2018slugnerds}, coreference resolution~\cite{chen2017robust,zhou2018they}, emotion detection~\cite{zahiri2018emotion}, and machine reading comprehension~\cite{ma-2018-challenging,sundream2018,yang2019friendsqa}. Besides, some pioneer studies focus on participating in dialogues~\cite{yoshino-2011-spoken,hixon-2015-learning} by asking users relation-related questions or using outputs of existing RE methods as inputs of other tasks~\cite{kluwer-2010-using,wang-cardie-2012}. In comparison, we focus on extracting relation triples from human-human dialogues, which is still under investigation.

\section{Conclusions}

We present the first human-annotated dialogue-based RE dataset DialogRE. We also design a new metric to evaluate the performance of RE methods in a conversational setting and argue that tracking speakers play a critical role in this task. We investigate the performance of several RE methods, and experimental results demonstrate that a speaker-aware extension on the best-performing model leads to substantial gains in both the standard and conversational settings. 

In the future, we are interested in investigating the generality of our defined schema for other comedies and different conversational registers, identifying the temporal intervals when relations are valid~\cite{surdeanu2013overview} in a dialogue, and joint dialogue-based information extraction as well as its potential combinations with multimodal signals from images, speech, and videos.

\section*{Acknowledgments}
We would like to thank the anonymous reviewers for their constructive comments and suggestions.

\bibliography{acl2020}
\bibliographystyle{acl_natbib}

\clearpage
\newpage

\appendix
\section{Appendices}
\label{sec:appendix}

\subsection{Definitions of New Relation Types}
\label{sec:newreldef}

We follow the original guideline to annotate relation types in the TAC-KBP SF task (marked with $\star$) unless stated otherwise and define new relation types as follows except for self-explainable types such as \textsc{per:major} and \textsc{per:client}. In this section, we keep the original speaker names in all examples for better readability.

\noindent $\circ$ \textbf{per:alternate\_names$^\star$}: Names used to refer a person that are distinct from speaker names or the first name mention in the given dialogue. It is possible to provide correct objects for this relation type without any contextual information such as triggers. Alternate names may include nicknames, first name, aliases, stage names, alternate transliterations, abbreviations, alternate spellings, full names, and birth names. However, if the full name mention appears first, we do not regard a first/last name alone as a valid value. An alternate name can also be a single word or a noun phrase.

\noindent $\circ$ \textbf{per:positive\_impression}: Have a positive impression (psychological) towards an object (\eg, a person, a book, a team, a song, a shop, or location). A named entity is expected here.

\noindent $\circ$ \textbf{per:negative\_impression}: Have a negative impression (psychological) towards an object. A named entity is expected here.

\noindent $\circ$ \textbf{per:acquaintance}: A person one knows slightly (\eg, name), but who is not a close friend.

\noindent $\circ$ \textbf{per:alumni}: Two persons studied in the same school, college, or university, not necessarily during the same period. Two persons can be in different majors. Classmates or batchmates also belong to this relation type.

\noindent $\circ$ \textbf{per:boss}: In most cases, we annotate B as the boss of A when A directly reports to B and is managed by B at work. In the meantime, A is the subordinate of B.
For example, we label (\emph{``Rachel''}, per:boss, \emph{``Joanna''}) and its corresponding trigger \emph{``assistant''} based on dialogue D1.   

\begin{table}[h!]
\centering
\footnotesize
\begin{tabular}{p{1cm}p{5.8cm}}
\toprule
\textbf{D1} &\\
Rachel: & Oh, uh, Joanna I was wondering if I could ask you something. There's an opening for an assistant buyer in Junior Miss...\\
Joanna: &  Okay, but that would actually be a big step down for me.  \\
Rachel: & Well, actually, I meant for me. The hiring committee is meeting people all day and... \\
Joanna: & Oh. Well, I wish I could say no, but you can’t stay my assistant forever. Neither can you Sophie, but for different reasons. \\
\bottomrule
\end{tabular}
\end{table}

\noindent $\circ$ \textbf{per:friends}: The subject cannot be imaginary. For example, \emph{``Maurice''} is not regarded as a friend of Joey based on dialogue D2.

\begin{table}[h!]
\centering
\footnotesize
\begin{tabular}{p{1cm}p{5.8cm}}
\toprule
\textbf{D2} &\\
Ross: & Joey had an imaginary childhood friend. His name was? \\
Monica: & Maurice.   \\
Ross: &  Correct, his profession was? \\
Rachel: &  Space cowboy! \\
\bottomrule
\end{tabular}
\end{table}

\noindent $\circ$ \textbf{per:girl/boyfriend}:  A relatively long-standing relationship compared to \textsc{per:positive\_impression} and \textsc{per:dates}, including but not limited to ex-relationships, partners, and engagement. The fact that two people dated for one or several times alone cannot guarantee that there exists a \textsc{per:girl/boyfriend} relation between them; we label \textsc{per:dates} for such an argument pair, instead.

\noindent $\circ$ \textbf{per:neighbor}: A neighbor could be a person who lives in your apartment building whether they are next door to you, or not. A neighbor could also be in the broader sense of a person who lives in your neighborhood. 

\noindent $\circ$ \textbf{per:roommate}: We regard that two persons are roommates if they share a living facility (\eg, an apartment or dormitory), and they are not family or romantically involved (\eg, per:spouse and per:girl/boyfriend).

\noindent $\circ$ \textbf{per:visited\_place}: A person visits a place in a relatively short term of period (\vs~\textsc{per:place\_of\_residence}). For example, we annotate (\emph{``Mike''}, per:visited\_place, \emph{``Barbados''}) in dialogue D3 and its corresponding trigger \emph{``coming to''}. 

\begin{table}[h!]
\centering
\footnotesize
\begin{tabular}{p{1cm}p{5.8cm}}
\toprule
\textbf{D3} &\\
Phoebe: & Okay, not a fan of the tough love. \\
Precious: & I just can't believe that Mike didn't give me any warning.  \\
Phoebe: &  But he didn't really know, you know. He wasn't planning on coming to Barbados and proposing to me... \\
Precious: &  He proposed to you? This is the worst birthday ever.\\
\bottomrule
\end{tabular}
\end{table}

\noindent $\circ$ \textbf{per:works}: The argument can be a piece of art, a song, a movie, a book, or a TV series.  

\noindent $\circ$ \textbf{per:place\_of\_work}: A location in the form of a string or a general noun phrase, where a person works such as \emph{``shop''}.

\noindent $\circ$ \textbf{per:pet}: We prefer to use named entities as arguments. If there is no name associated with a pet, we keep its species (\eg, dog) mentioned in a dialogue.

\subsection{Relation Type Distribution}
\label{sec:reltypedist}

\subsection{Distance Between Argument Pairs}
\label{sec:argdist}

\begin{figure}[th!]
   \begin{center}
   \includegraphics[width=0.42\textwidth]{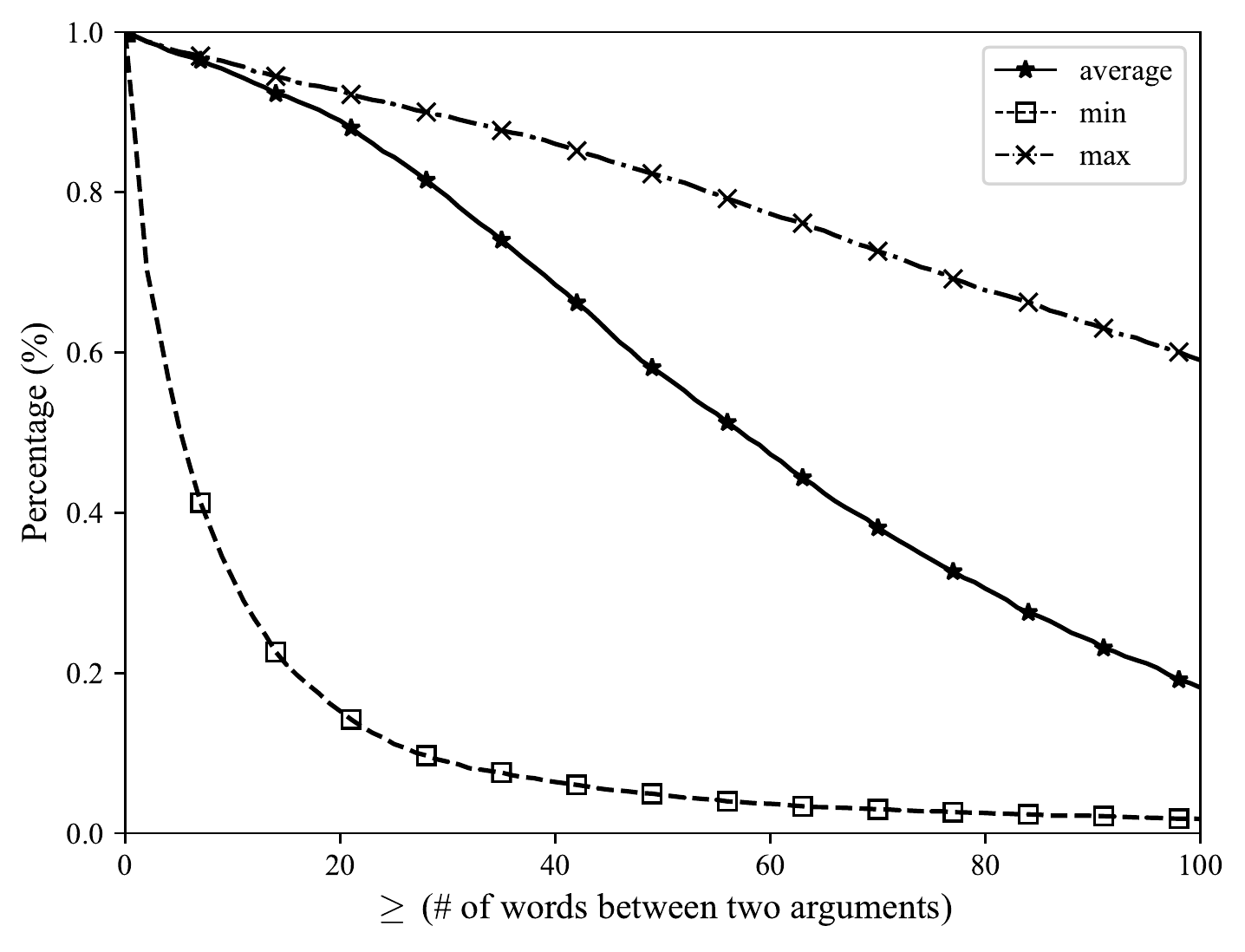}
   \end{center}
\caption{Number of words between two arguments within a dialogue in DialogRE.} 
\label{fig:distance}
\end{figure}

\subsection{Other Input Sequences}
\label{sec:input_sequence}
We also experiment with the following three alternative input sequences on the BERT baseline: (1) \texttt{[CLS]}$d^\#$\texttt{[SEP]}, (2) \texttt{[CLS]}$d^\#$\texttt{[SEP]}$a_1$\texttt{[SEP]}$a_2$\texttt{[SEP]}, and (3) \texttt{[CLS]}$d''$\texttt{[SEP]}, where $d^\#$ is obtained by replacing subject/object mentions in $d$ with special tokens \texttt{[SUBJ]} and \texttt{[OBJ]}, and $d''$ is obtained by surrounding each mention of $a_i$ ($i\in\{1,2\}$) in $d$ with special tokens \texttt{[A\textsubscript{i}]} and \texttt{[/A\textsubscript{i}]}~\cite{baldini2019matching}. The F$1$ of them is $50.9\%$,  $58.8\%$, and $57.9\%$, respectively, substantially lower than that of $\text{BERT}_\text{S}$ ($61.2\%$). %

\subsection{Ground Truth Argument Type}
\label{sec:gtargtype}

\noindent \textbf{Method 3} follows the input sequence employed by \citet{joshi2019spanbert}. Specifically, we replace the argument mentions in document $d$ with newly-introduced special tokens that represent the subject/object and argument types. For example, if the subject type is PER and the object is STRING, we replace every subject mention in $d$ with \texttt{[SUBJ-PER]} and every object mention with \texttt{[OBJ-STRING]}. Let $d'$ denote the new document. The input sequence is \texttt{[CLS]}$d'$\texttt{[SEP]}.
    
\noindent \textbf{Method 4} takes as input the sequence employed by \citet{shi2019simple}. The input sequence is \texttt{[CLS]}$d'$\texttt{[SEP]}$a_1$\texttt{[SEP]}$a_2$\texttt{[SEP]}, where $d'$ is defined in Method 3. %

\end{document}